\documentclass[article]{jss}

\usepackage{lmodern}
\usepackage{amsmath}
\usepackage{algorithm}
\usepackage[noend]{algpseudocode}
\usepackage{multirow}
\usepackage{framed}
\usepackage{listings}
\usepackage{amsfonts}

\author{Neha R. Gupta\\ Duke University
\And Vittorio Orlandi\\ Duke University
\And Chia-Rui Chang\\ Harvard University
\AND Tianyu Wang\\ Fudan University
\And Marco Morucci\\ New York University
\And Pritam Dey\\ Duke University
\AND Thomas J. Howell\\ Duke University
\And Xian Sun\\ Duke University
\And Angikar Ghosal\\ Duke University
\AND Sudeepa Roy\\Duke University
\And Cynthia Rudin\\Duke University
\And Alexander Volfovsky\\Duke University}
\Plainauthor{Neha Gupta}

\title{\pkg{dame-flame}: A \proglang{Python} Library Providing Fast Interpretable Matching for Causal Inference}
\Plaintitle{\pkg{dame-flame}: A Python Library Providing Fast Interpretable Matching for Causal Inference}
\Shorttitle{\pkg{dame-flame}: Fast Interpretable Matching for Causal Inference}

\Abstract{
\pkg{dame-flame} is a \proglang{Python} package for performing \emph{matching} for \emph{observational causal inference} on datasets containing discrete covariates. This package implements the \emph{dynamic almost matching exactly (DAME)} and \emph{fast, large-scale almost matching exactly (FLAME)} algorithms, which match treatment and control units on subsets of the covariates. The resulting matched groups are interpretable, because the matches are made directly on covariates, and high-quality, because machine learning is used to determine which covariates are important to match on instead of human inputs. The package provides several adjustable parameters to adapt the algorithms to specific applications, and can calculate treatment effects after matching. The most recent source code of the implementation is available at https://github.com/almost-matching-exactly/DAME-FLAME-Python-Package
}

\Plainkeywords{Causal inference, machine learning, matching, Python}

\Address{
Neha R. Gupta\\
Duke University\\
Durham, North Carolina\\
Email: \email{neha.r.gupta@duke.edu}\\

Sudeepa Roy\\
Department of Computer Science\\
Duke University\\
Durham, North Carolina\\
E-mail: \email{sudeepa@cs.duke.edu}\\
\\
Cynthia Rudin\\
Department of Computer Science, Department of Electrical and Computer Engineering \\
Duke University\\
Durham, North Carolina\\
E-mail: \email{cynthia@cs.duke.edu}\\
\\
Alexander Volfovsky\\
Department of Statistical Science\\
Duke University\\
Durham, North Carolina\\
E-mail: \email{alexander.volfovsky@duke.edu}\\
}

\begin{document}



\pagebreak

\section{Introduction} \label{sec:intro}

The \pkg{dame-flame} \proglang{Python} package is the first major implementation of two algorithms, the \emph{dynamic almost matching exactly} (DAME) algorithm (\citealt{liu2018interpretable}, published in AISTATS'19), and the \emph{fast, large-scale almost matching exactly} (FLAME) algorithm (\citealt{wang2017flame}, published in JMLR'21), which provide \textit{almost exact} matching of treatment and control units in discrete observational data for causal analysis. As discussed in \cite{liu2018interpretable}, and \cite{wang2017flame}, the two algorithms produce high-quality interpretable matched groups, by using machine learning on a holdout training set to learn distance metrics. DAME solves an optimization problem that matches units on as many covariates as possible, prioritizing matches on important covariates. FLAME approximates the solution found by DAME via a much faster backward feature selection procedure.

 The DAME and FLAME algorithms are discussed in the remainder of this section. We also provide testing and installation details. In Section \ref{sec:codeexplanation}, we discuss the class structure in the \pkg{dame-flame} package, detail special features of \pkg{dame-flame}, and compare \pkg{dame-flame} to other matching packages. In Section \ref{sec:examples}, we offer examples and a user guide. 
 
\subsection{Algorithms overview} 
\label{sec:algosummary}
 
The advantage of matching is that it accounts for confounding of treatment effect estimates, and permits interpretable analyses that are easier to troubleshoot than other types of analysis for observational causal studies. However, matching is not trivial; in high dimensional settings, few individuals can be matched exactly on all covariates, so other ways must be found. We offer the \proglang{Python} package \pkg{dame-flame} to support \emph{almost exact matching} in a way that focuses on identifying important subsets of the covariates using machine learning and matching units exactly on those subsets.

\pkg{dame-flame} is designed for causal inference problems with a binary treatment variable, an observed outcome variable, and any number of pre-treatment covariates. Several assumptions standard in observational causal inference must be made in order for the DAME and FLAME algorithms to be applicable. One is the stable unit treatment value assumption (SUTVA), which assumes that treatments applied to one unit do not affect the outcome of other units, and there is only one version of treatment. A second requirement is that of unconfoundedness, or ignorability. It is important that the outcome is independent of the treatment assignment. A final requirement is overlap of treatment and control groups. The treatment and control groups are said to not have any overlap at some location in a distribution when the probability of being treated at that location is either exactly 0 or 1. If there is no overlap for all covariates, then FLAME and DAME algorithms would not be able to find any matches.

A more moderate issue is when only few treated and control units overlap in covariate values, or partial overlap, where we may not find both treatment and control units with sufficient overlap to match with.
In this case, the user's settings on the \pkg{dame-flame} package would determine what \emph{quality of matches} would be acceptable. Match quality is discussed further in Section \ref{sec:algomethodology}.

Units that were not able to be matched are not included in treatment effect calculations. 

Discrete observed covariate data is a requirement of \pkg{dame-flame}. We do not recommend that users bin continuous data, with an exception for scenarios in which users are confident they are binning variables in a way that is meaningful for their research. 

\pkg{dame-flame} is efficient, owing to a combination of fast bit-vector computations, and a backwards feature elimination process (for FLAME) or a type of downwards closure property for systematic feature elimination (for DAME). Therefore, FLAME is faster, but DAME is able to match units on more covariates. 

We also support a hybrid execution of FLAME and DAME methods. 
The combination of FLAME (at earlier iterations) and DAME (at later iterations) permits faster elimination of irrelevant covariates in the earlier iterations and then a more careful elimination of covariates in the later iterations, thereby achieving a trade-off between scalability and quality.

\subsection{Algorithm methodology}
\label{sec:algomethodology}

In this section we describe the algorithms implemented. First we discuss the mathematical problem that FLAME and DAME aim to solve, then we describe the steps of each algorithm.

Suppose we have $n$ units, indexed by $i$, and $p$ covariates. We may interchangeably refer to units as `individuals' or `observations'. 
Formally, consider a dataframe $D = [X,Y,T]$, including $n \times p$ matrix $X\in \{0,1,\dots,k\}^{n\times p}$ where 
$X$ contains the categorical 
covariates for all units, 
$Y\in \mathbb{R}^n$ denotes the outcome vector, and $T\in\{0,1\}^n$ denotes the treatment indicator vector
($1$ for treated, $0$ for control); ${\mathbf x}_i$ denotes the covariate vector of unit $i$.  

We will use $\theta\in\{0,1\}^p$ to denote the variable selection indicator vector for a subset of covariates to match on. A unit is a triplet (covariate value $\mathbf x_i$, observed outcome $y_i$, treatment indicator $t_i$). Given dataset $\mathcal{S}$, define the \textit{matched group} for unit $i$ with respect to covariates selected by $\theta$ as the units in $\mathcal{S}$ that match $i$ exactly on  covariates $\theta$:
$$
\mathtt{MG}_i (\theta, \mathcal{S} ) = \{i^{'} \in \mathcal{S} : \mathbf x_{i^{'}} \circ \theta = \mathbf x_i \circ\theta \},
$$
where $\circ$ denotes Hadamard product. Under the assumption of no unobserved confounding
the question of the causal effect of $T$ on $Y$ then becomes which covariates $\theta$ we should match unit $i$ on. 

In FLAME and DAME, the value of a set of covariates $\theta$ is determined by how well these covariates can be used together to predict outcomes. However, we often prefer not to look at the outcomes of our dataset to determine how to match, to avoid risk of biasing the estimates. Thus, we consider a separate training dataset $\mathcal{S}^{tr}$.
Let $\mathcal{S}_0^{tr}$ be the subset (of $\mathcal{S}^{tr}$) of control units ($(X^{tr},Y^{tr})$ with $T^{tr} = 0$), and let $\mathcal{S}_1^{tr}$ be the subset (of $\mathcal{S}^{tr}$) of treated units ($(X^{tr},Y^{tr})$ with $T^{tr} = 1$). The empirical prediction error $\hat{\mathtt{PE}}_{\mathcal{F}_{ \left\|\theta \right\|_0}} $ is defined with respect to a class of functions $\mathcal{F}_k := \{ f: \{0,1\}^k \rightarrow [0,1] \}$ ($1 \le k \le d$) as: 
\begin{equation}
\begin{split}
\hat{\mathtt{PE}}_{\mathcal{F}_{ \left\|\theta \right\|_0}} (\theta, \mathcal{S}^{tr})
=
\min_{f^{(1)} \in \mathcal{F}_{ \left\|\theta \right\|_0}} \frac{1}{ | \mathcal{S}_1^{tr} |}
\sum_{ (\mathbf{x}_i, y_i) \in \mathcal{S}_1^{tr} } (f^{(1)}(\mathbf{x}_i^{} \circ \theta )-y_i)^2\nonumber
 + \\
\min_{ f^{(0)} \in \mathcal{F}_{ \left\|\theta \right\|_0}} \frac{1}{ | \mathcal{S}_0^{tr} |} 
\sum_{ (\mathbf{x}_i, y_i) \in \mathcal{S}_0^{tr} }
(f^{(0)}(\mathbf{x}_i^{} \circ \theta )-y_i)^2. 
\end{split}
\end{equation} 
That is, $\hat{\mathtt{PE}}_{\mathcal{F}_{ \left\|\theta \right\|_0}} $ is the smallest prediction error we can get on both treatment and control populations using the features specified by $\theta$. Thus, given a matching dataset $\mathcal{S}^{ma}$ and a training dataset  $\mathcal{S}^{tr}$, the best selection indicator we could achieve for a nontrivial matched group that contains treatment unit $i$ would be: 
\begin{eqnarray}\label{opt:flame}
\theta_{i, \mathcal{S}^{ma}}^{*} \in \mathrm{arg} \min_{\theta} \hat{\mathtt{PE}}_{\mathcal{F}_{ \left\|\theta \right\|_0} } (\theta, \mathcal{S}^{tr})  \textrm{ s.t. }\exists \ell\in \texttt{MG}_i ({\theta}, \mathcal{S}^{ma}) \textrm{ s.t. } t_{\ell}=0.
\end{eqnarray}
This constraint says that the matched group contains at least one control unit. It also matches on covariates that together can be used to predict well on the training set. The covariates selected by $\theta_{i, \mathcal{S}^{ma}}^{*}$ are those that predict the outcome best, provided that at least one control unit has the same exact covariate values as $i$ on the covariates selected by  $\theta_{i, \mathcal{S}^{ma}}^{*}$. Please note that the predictive error is not the sole determinant of a matched group, and the covariates used in a matched group is determined based on an iterative procedure. This is discussed in more detail below.


The \textit{main matched group} for $i$ is then defined as $\mathtt{MG}_i ( \theta^*_{i, \mathcal{S}^{ma}}, \mathcal{S}^{ma})$.
Users can choose whether units are matched with replacement; that is, whether a previously matched unit can be matched in a subsequent iteration of the algorithm. The first time a unit is matched, that matched set is its \textit{main matched group}, from which its treatment effect estimates are calculated. If units are allowed to be matched with replacement, a unit can become a member of another unit's main matched group. Any additional groups which a unit belongs to other than its \textit{main matched group} is its \textit{auxiliary matched group}. 

The goal of FLAME and DAME is to calculate the main matched group ${\mathtt{MG}}_i (\theta^*_{i, \mathcal{S}^{ma}},  \mathcal{S}^{ma})$ \textit{for as many units $i$ as possible.} Any units without a main matched group are likely outliers that cannot be easily matched. Then, the matched groups can be used to estimate treatment effects.

\begin{figure}[h!]
  \centering
  \includegraphics[width=\textwidth]{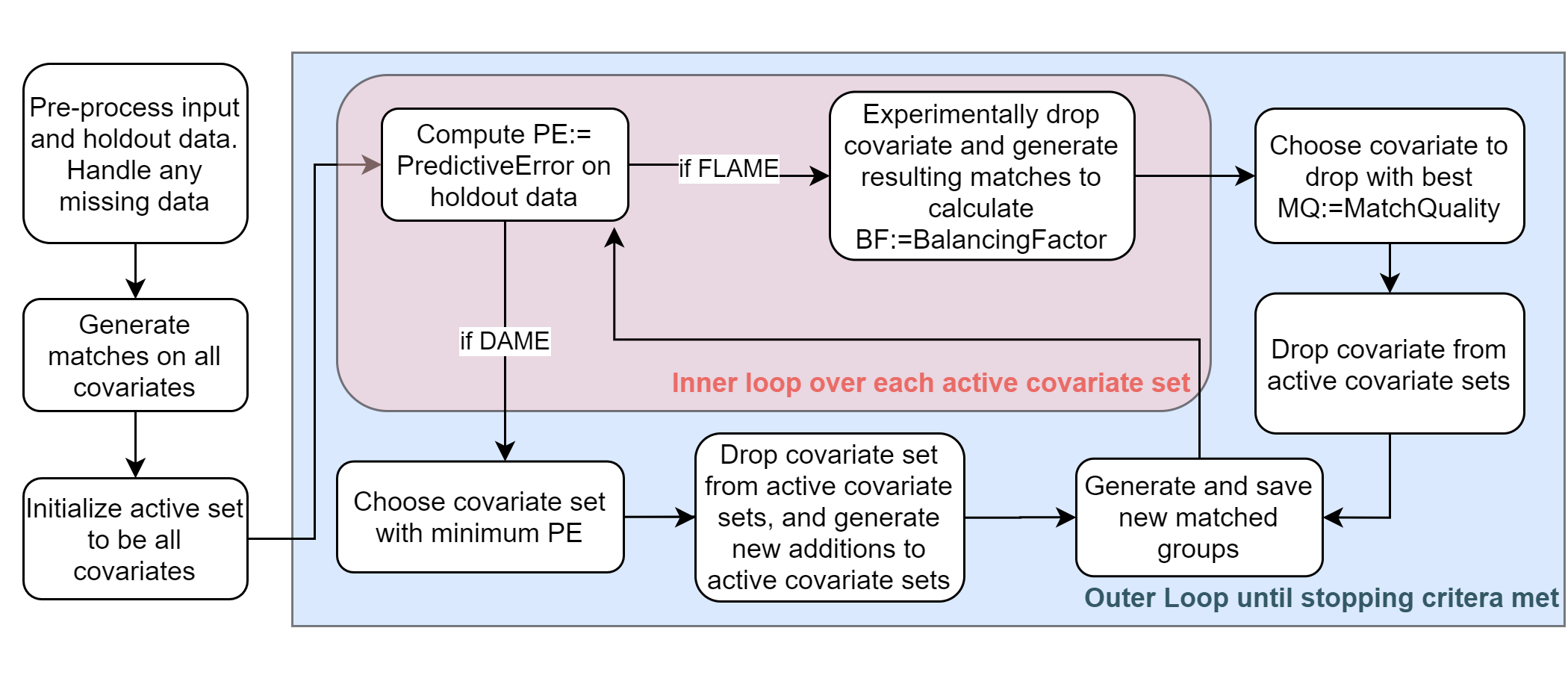}
  \caption{Flow diagram: DAME/FLAME algorithms}
  \label{fig:flowdiagram}
\end{figure}

The implementation of the above mathematical descriptions in both DAME and FLAME algorithms requires us to iterate over two nested loops, shown in Figure \ref{fig:flowdiagram}.
 
Users must begin by providing a dataset with discrete observational covariates, a binary treatment indicator, and a continuous or discrete outcome column. If users do not supply a separate holdout training dataset with the same covariates as the input dataset, optional parameters of the package, discussed in further detail in Section \ref{sec:user-guide}, will allow them to partition the input dataset to create the holdout training set. 

Users have a variety of options for handling {\em missing covariate data}. They can (1) exclude units with missing values from the procedure, (2) impute the missing data via the  multiple imputation by chained equations (MICE) method \citep{buuren2010mice}, or (3) specify that matches should not occur on missing values without imputing them. In this third case, matches can still be made for a unit on its covariates that are not missing. 

As described above, after handling missing data, the algorithms begin by matching any units that can be matched exactly on all covariates, where at least one treatment unit as well as at least one control unit are contained in each matched group. The algorithms then execute the outer loop: updating sets of covariates to match on, referred to as \textit{active covariate sets}, until a stopping criterion is reached. In each iteration, the algorithms execute the inner loop, examining each covariate set to select the best one to match on. Units that have identical values for all of the covariates that are part of the chosen covariate set form a matched group, as long as at least one treatment unit and at least one control unit are present in the group.

To determine the best covariate set, FLAME selects the covariates yielding the highest \emph{match quality} $\mathtt{MQ}$, defined as  $\mathtt{MQ} = \mathtt{C} \cdot \mathtt{BF} - \mathtt{PE}$, where $\mathtt{C}$ is a user-specified hyperparameter. DAME selects the covariate set minimizing $\mathtt{PE}$. Here, $\mathtt{PE}$ denotes the \emph{predictive error} as described above. The \emph{balancing factor}, $\mathtt{BF}$, measures the proportion of treatment and control units that are matched on a covariate set. DAME iterates efficiently over covariate sets, prioritizing matching on large covariate sets if they can be used to effectively predict the outcome on the holdout training set. FLAME approximates this solution for scalability: it consistently matches on a smaller set of covariates than in the previous iteration, while still ensuring that each covariate set can be used to effectively predict the outcome on the holdout training set. It is important to note that $\mathtt{MQ}$ can be replaced by any non-adaptive, pre-determined, measure of quality that a user might be interested in. 

The outer loop has a number of possible stopping criteria. It must stop when all units are placed in matched groups or all covariate sets have been dropped. Additionally, users can enforce stopping based off other criteria, e.g., (1) when there are too few unmatched (treatment or control) units, (2) after a certain number of iterations, (3) when predictive error rises too much, or (4) when the balancing factor for a given round is not high enough. If the third criteria is chosen (when predictive error rises too much), then any units that have already been matched were matched on a set of covariates that together can be used to predict the outcome well.

\subsection{Installation, setup, and testing}
 The package is designed for \proglang{Python} 3.6 and above.
\pkg{dame-flame} depends on \pkg{scikit-learn} version 0.23.2 and above, \pkg{pandas} version 0.11.0 and above, and \pkg{numpy} version 1.16.5 \citep{scikit-learn, reback2020pandas, numpy, mckinney-proc-scipy-2010}.
 
 \pkg{dame-flame} is available for download on PyPi and on GitHub\footnote{
Code is publicly available at:\\ \url{https://github.com/almost-matching-exactly/DAME-FLAME-Python-Package/}.}. The public documentation website\footnote{\url{https://almost-matching-exactly.github.io/DAME-FLAME-Python-Package/}} covers the API, installation instructions, a quick-start tutorial, several examples, and a contributing guide. In harmony with the best open source practices, users are invited to report any unexpected bugs, assist with cleaning or maintain code, add details or use-cases to the documentation, and add more test cases. They are welcome to do so via the GitHub bug reporting and pull requesting features, or by directly emailing the core development and research team. 

There is also an accompanying \proglang{R} package for the FLAME and DAME algorithms, that can also be found on GitHub\footnote{\url{https://github.com/almost-matching-exactly/R-FLAME}}. The \proglang{R} package is also an open source package welcoming user questions and contributions.

Testing was done to ensure that \pkg{R-FLAME} and the \pkg{dame-flame} \proglang{Python} package yield consistent results on a range of datasets and parameter options.

For further reliability testing, \pkg{dame-flame} offers continuous integration through Travis-CI, and the independent Coveralls API was used to verify the test suite offers an extensive code coverage. 
 
 \section{Code and its explanation}\label{sec:codeexplanation}
 
 \subsection{Class structure and advanced features}
The API consists of standard Pythonic design established by \pkg{scikit-learn}. The main feature of the API is the class \code{matching.MatchingParent}, and two subclasses \code{DAME} and \code{FLAME}. These offer standard methods \code{fit}, where a user provides a holdout training dataset, and \code{predict}, where a user provides a matching dataset, and the matches of the algorithm are computed.

Continuing to follow \pkg{scikit-learn}'s standards, any post-processing is done using \code{utils} functions, which take as arguments a \code{matching.DAME} or \code{matching.FLAME} object, and use these to compute matched groups of units, and estimate treatment effects, including the \emph{average treatment effect} (ATE) for the population, and \emph{conditional average treatment effect} (CATE) of a selected unit. These are mathematically defined in Section \ref{sec:def-estimators}. 

In the rest of this section, we proceed to discuss advanced features for computing predictive error, early stopping features, missing data handling options, and FLAME specific options. For each of these advanced features, we discuss the theory, list the specific parameter for users to select, and the package default for this feature. We conclude this section by comparing these features to features of other popular matching packages. 

\subsection{Definitions of estimands and estimators} \label{sec:def-estimators}
We continue with notation previously introduced: Units are indexed by $i$, which ranges from 1 to $N$. We may interchangeably refer to units as `individuals' or `observations'. Also, note that all units which contribute to treatment effect estimates must have  been matched.

There are $p$ pre-treatment covariates $x_1, \dots, x_p$ and for a given unit $i$, we will refer to its vector of covariates as $X_i$. Let the binary treatment indicator for unit $i$ be denoted by $T_i$. We let $Y_i$ be the observed outcome for individual $i$ where $Y_i = Y_i(1)T_i + Y_i(0)(1 - T_i)$ and $Y_i(0), Y_i(1)$ are the potential outcomes of unit $i$ under control and treatment, respectively. Lastly, we introduce notation for matched groups, which we index by $m$, which ranges from 1 to $M$. The size of a matched group $m$ is $\|m\|$, which is the number of units in the group.

The conditional average treatment effect, or CATE, is defined as the average treatment effect conditional on particular covariates. Formally, given a set of covariates $X_i$, the CATE is: 
\[\text{CATE}(X_i)=\frac{1}{N}\sum_{i=1}^N\mathbb{E}[Y(1)-Y(0)\|X_i].\]
Our implementation of CATE estimation allows users to input a unit $i$ and receive its CATE estimate, based on its main matched group. 

Since our units are matched almost-exactly, all units sharing the same main matched group will have the same CATE estimate. For a unit $i$ whose main matched group $m$ is of size $\|m\|$, we estimate its CATE as: $$\frac{1}{\|m\|}\sum_{i:T_i=1}[\hat{Y}_i(1)]-\frac{1}{\|m\|}\sum_{i:T_i=0}[\hat{Y}_i(0)]$$ 
where $\hat{Y}_i(0), \hat{Y}_i(1)$ are calculated using the units in $m$ with treatment $1 - T_i$.

The Average Treatment Effect (ATE) is unconditional on X: $\text{ATE} = \mathbb{E}[Y(1)-Y(0)]$. We offer two estimators of the ATE. For one of the estimators, we will create a weighted combination of CATE estimates from the various matched groups.
Let $q_i$ denote the number of matched groups that unit $i$ appears in. Note this quantity can be greater than 1 when matching is done with replacement, via the \code{repeats} argument. We then define the weight of a matched group $m$ as $w_m=\sum_{i=1}^{\|m\|}\frac{1}{q_i}$. We denote the CATE estimate of group $m$ as $\text{{CATE}}_m$. 
We estimate the ATE as: $$\text{ATE} = \frac{\sum_{m}\text{CATE}_m \times w_m}{\sum_{m}w_m}.$$
Note that this expression downweights units that were matched many times so that they do not dominate the ATE estimate.

The second ATE estimator implemented is the simple matching estimator described in \cite{abadie}. We also offer the variance estimator of this ATE estimate provided by \citep{abadie}. Please note that this estimator assumes constant treatment effects and homoscedasticity. It is not asymptotically normal, so users must make the standard implications on confidence intervals or hypothesis tests based on this.

\subsection{Predictive error}
As discussed in Section \ref{sec:intro}, the algorithm's decision of best covariates to match on relies on a computation of predictive error, or PE, based on a user-chosen machine learning algorithm run on the holdout dataset. The \pkg{dame-flame} \proglang{Python} package offers different options for the machine learning algorithm used, as well as a simplified FLAME and simplified DAME that does not use machine learning, but instead allows the users to input feature importance information for matching. We use \pkg{scikit-learn} for the underlying learning algorithms, and refer the reader to their documentation and references to learn more about these popular machine learning algorithms, as well as their specific implementations, applied separately to the treated and control units in the holdout set.

Users can easily modify the code to feature a new machine learning algorithm of their choice. The options we provide at this time for the computation of PE include the following. 
\begin{itemize}
    \item Ridge Regression. 
    Ridge regression minimizes a residual sum of squares plus a regularization term measuring the $\ell_2$ norm of the coefficient vector, multiplied by a shrinkage parameter, $\alpha$. For this option, a larger $\alpha$ should be chosen if it is believed that there is greater multicollinearity in the data, meaning that many covariates are linearly correlated. This option can be chosen using the parameter \code{adaptive_weights=`ridge'}. The $\alpha$ parameter can also be adjusted using parameter \code{alpha} when declaring a matching object. 
    \item Cross-Validated Ridge Regression. This is a ridge regression with built-in cross validation to determine the best $\alpha$ parameter. We use the \pkg{scikit-learn} \code{ridgeCV} class, but the default array of $\alpha$ options that we provide the function to iterate over is larger than the default they provide, for greater flexibility.
    This option is advantageous over the ‘ridge’ option without cross validation in the case when a user is uncertain about the $\alpha$ parameter, and a minor speed decrease owing to cross validation is acceptable. This option can be chosen using the parameter \code{adaptive_weights=`ridgeCV'}.
    \item Decision Tree. Designed on a variation of the CART algorithm, this is the only option that can be used for unordered discrete data. This option can be chosen using the parameter \code{adaptive_weights=`decision-tree'} as described in Section \ref{sec:user-guide}.
\end{itemize}

The option a user chooses can be selected using the specified value for the parameter \linebreak \code{adaptive_weights} when declaring a matching object, as shown in examples  in Section \ref{sec:examples}. If, instead of allowing the algorithm to select covariates via the PE parameter, the user prefers to pre-specify covariate importance, they can do so by specifying \code{adaptive_weights = False}. The weights to the covariates in \code{input_data} can be specified using the parameter \code{weight_array} in the \code{fit} function. The values in that array must sum to 1.

\subsection{Early-stopping options} \label{sec:early-stopping} The FLAME and DAME algorithms will stop after running to completion, or based on user-defined early stopping criteria. The default option is that the algorithm runs until all units are matched, or until there is a large spike in predictive error. If runtime or high accuracy of estimates of treatment effects are important, then we recommend users experiment with their stopping criteria based on their specific needs and dataset size. A large dataset will have a longer runtime, and an early stop will take less time. Without early stopping, the matches could degrade in quality in later iterations, where units that are farther from each other in covariate space would now be matched, leading to worse overall performance of the method.

Below, we define and discuss the early stopping criteria that users can choose. All criteria are controlled via a parameter to the classes defined in Section \ref{sec:user-guide}. 

\begin{itemize}
    \item The maximum number of iterations of the FLAME or DAME algorithm, via the parameter \code{early_stop_iterations}. Iterations start at 0 so that a value of 0 leads to only exact matches being made. If FLAME is used, then this is the maximum number of covariates that can be dropped, meaning when the total number of covariates is $m$, no unit will be matched on fewer than $m-$\code{early_stop_iterations} covariates. This is useful when the user wants only matches of a specific high level of quality, or when the user is concerned about computational time.
    \item Unmatched units in treatment or control, via the parameters \code{stop_unmatched_c} and \code{stop_unmatched_t}. When the algorithm is set with the \code{repeats=True} parameter, then previously matched units (that is, units whose main matched groups have already been determined) can still be placed in the main matched groups of other units. The algorithm will by default stop iterating when there are no more units that have not been placed in any group. 
    \item Proportion of unmatched units, via the parameters \code{early_stop_un_c_frac}, and \\ \code{early_stop_un_t_frac}. This stops the algorithm when the fraction of control units or treatment units are unmatched goes below a user-defined value. One specific case in which this could be useful is where a user thinks that some percent of the input is unlikely to result in good matches.	
    \item Predictive error, via the parameter \code{early_stop_pe}. The predictive error measures how important a covariate set is for predicting the outcome on the holdout training dataset, using a machine learning algorithm. It is the sole determiner of the covariate set to match on for DAME, and one of two factors for FLAME. If FLAME or DAME attempts to drop a covariate set that would raise the predictive error above (1 + \code{early_stop_pe}) times the baseline predictive error (the predictive error when using all covariates), the algorithm terminates without dropping this covariate set. 
\end{itemize}

\subsection{Missing data handling options} Users are offered a variety of options for handling missing covariate data. Imputing missing values in datasets is possible, but matches become less interpretable when matching on imputed values, in that it is more difficult to discern why a match was recommended by the matching algorithm. Here, we discuss the options we provide in detail and make recommendations. The parameter to select in the \pkg{dame-flame} \proglang{Python} package is mentioned here, and more details on usage is provided in Section \ref{sec:user-guide}.

There can be missing data in either the input matching data, the holdout training data, or both. The specific character that is used to denote missing value can be selected via the parameter \code{missing_indicator}, which can be a character, integer, or numpy.nan.

For the input dataset, three options exist: 
\begin{itemize}
    \item Omit units with missing values. We recommend using this if missing values indicate data fidelity issues in a unit. The algorithms handle this by ensuring that units in the input dataset that have missing data are dropped from the dataset prior to running the algorithms finding the matches. This option is selected via the parameter \code{missing_data_replace=1}.
    \item Match units with missing values, but ignore missing values when considering which units to match to. We recommend this for the majority of cases. The underlying algorithm will handle this when pre-processing the input. 
    This option is selected via the parameter \code{missing_data_replace=2}.
    \item Impute missing values with MICE. This is computationally costly and would reduce the interpretability of the matches. The algorithm would create several imputed datasets and iterate over each to find a match according to each dataset. This option is selected via the parameter \\\code{missing_data_replace=3}. The number of MICE imputations is selected via the parameter \code{missing_data_imputations}.
\end{itemize}

For the holdout dataset, the following two options exist:
\begin{itemize}
    \item Omit units that have any missing values. We recommend this option only if a missing completely at random assumption is tenable in both holdout and matching datasets \citep{mcar}. In the underlying algorithm, units in the holdout dataset that have missing data are dropped from the dataset prior to running the DAME or FLAME algorithm to find the matches. This option is chosen by using the parameter \linebreak \code{missing_holdout_replace=1}.
    \item Impute missing values with MICE. In the underlying algorithm, we begin by running MICE to create several imputed training holdout datasets. The DAME or FLAME algorithm is run once, and the best covariate set is chosen based on the predictive error over all imputed datasets. This option is chosen by using the parameter \linebreak \code{missing_holdout_replace=2}.
\end{itemize}

The underlying MICE implementation is done using \pkg{scikit-learn}’s experimental \pkg{IterativeImpute} package, and relies on Decision tree regressions in the imputation process, to ensure that the data generated is fit for unordered categorical data.

\subsection{Additional parameters available}
As discussed in Section \ref{sec:codeexplanation}, users can adjust whether they match units with or without replacement. This is controlled via the boolean parameter \code{repeats}.

Output style can also be controlled by the user, via a range of parameters. All of these parameters are used when declaring a matching object. 
\begin{itemize}
    \item The parameter \code{verbose}. This is a number that will range from 0 to 3 and higher numbers result in additional information being output. If true, the output of the algorithm will include the predictive error of the covariate sets used for matching in each iteration.
    \item The boolean parameter \code{want_pe}. If true, the output will include the predictive error for each iteration.
    \item  The boolean parameter \code{want_bf}. If true, the output will include the balancing factor for each iteration.
\end{itemize}

There are two FLAME specific parameters, which users would provide in the final \code{predict} step. These are:
\begin{itemize}
    \item \code{C}, type float. This is the tradeoff parameter between the balancing factor and the predictive error when deciding which covariates to match on.
    \item \code{pre_dame}, type \{float, integer\}, default=\code{float(`inf')}. The number of iterations to run FLAME before switching to DAME, allowing for a hybrid FLAME-DAME option.
\end{itemize}

\subsection{Comparison to other matching packages}

Many other matching methods either 
produce low-quality matches (leading to potentially poor treatment effect estimates), uninterpretable matches (e.g., in which matches include units with highly dissimilar covariates values), or matches that are manually defined by an analyst. 

One of the most widely used algorithms is nearest neighbor propensity score matching, provided by the \proglang{R} package \pkg{MatchIt} \citep{stuart2011matchit}. 
Propensity score matching reduces units' covariate information to one dimension, allowing matches to contain units even at extreme ends of the covariate space; such matches are uninterpretable. MatchIt allows other matching metrics, such as Mahalanobis distance, but does not allow for learning the proper metric as FLAME and DAME.

Another common matching algorithm is Coarsened Exact Matching (CEM), popularly available in the \proglang{R} package \pkg{cem} \citep{iacus2009cem}. CEM requires the user to manually coarsen variables, requiring humans to know detailed information about a high-dimensional space in advance, a task at which humans are not naturally adept  \citep{liu2018interpretable}. Coarsening covariates via default histogram binning methods fails to take into consideration their impacts on treatment and outcome, resulting in poor matches. Instead of requiring a human to manually input how matches should be constructed (or to use histogram binning), FLAME and DAME use machine learning on a training set to determine this information. 

Many of the features described in Section \ref{sec:early-stopping} and Section \ref{sec:algosummary} are unavailable in other matching packages. Table \ref{tab:comparison} compares the characteristics of \pkg{dame-flame} against popular alternatives. Most matching packages are implemented in \proglang{R}. \proglang{R}'s \pkg{cem} package only supports Average Treatment Effect on the Treated (ATT) treatment effects \citep{iacus2009cem}. 
The \pkg{MatchIt} package focuses on estimation of average treatment effects and not conditional average treatment effects, both of which are handled in the same coherent manner by \pkg{dame-flame}.
Users of any propensity score matching algorithm can adjust matched group sizes only by entering a ratio of treatment to control units, forcing all matched groups to be of the same size.  \proglang{Python}'s \pkg{PyMatch} and \pkg{DoWhy} offer propensity score matching \citep{pymatch}, but \pkg{DoWhy} does not emphasize matched groups, favoring to present treatment effects and other output. \pkg{DoWhy} uses the \pkg{EconML} package to provide conditional average treatment effect estimates \citep{dowhy, econml}.
\proglang{R}'s \pkg{cem} package is a good choice for datasets with multi-level, non-binary treatment variables, whereas the current version of \pkg{dame-flame} does not yet offer a multi-level treatment solution. 

\begin{table}
\begin{center}
\begin{tabular}{{|p{0.25\linewidth}|p{0.23\linewidth}|p{0.22\linewidth}|p{0.1\linewidth}|p{0.3\linewidth}|}}%
\hline
Language: Package & Built-in treatment-effect estimations & Missing data handling options & Provide matched groups     \\\hline
\proglang{Python}: \pkg{dame-flame}    & Average, conditional     & \checkmark        &  \checkmark        \\\hline
\proglang{Python}: \pkg{DoWhy}    & Average    &     &   \\\hline
\proglang{Python}: \pkg{PyMatch}    &     &      &   \checkmark     \\\hline
\proglang{R}: \pkg{cem}    &  Average    &  \checkmark     &  \checkmark       \\\hline
\proglang{R}: \pkg{MatchIt}{ Propensity Score }    & Average   &   \checkmark   &      \checkmark   \\\hline

\end{tabular}
\caption{{Features of matching packages.}}
\vspace{-3mm}
\label{tab:comparison}
\end{center}
\end{table}

A further advantage of \pkg{dame-flame} is the higher quality of the matched groups generated by DAME and FLAME relative to propensity score matching, as shown by \cite{liu2018interpretable}. 

A drawback of \pkg{dame-flame} is the requirement that covariates be discrete. The packages \pkg{DoWhy}, \pkg{PyMatch}, \pkg{cem} and \pkg{MatchIt} do allow users to use continuous covariates without any pre-processing steps or manual binning. Although a user could manually bin continuous covariate values prior to  using \pkg{dame-flame}, we do not recommend this asides scenarios in which users are confident they are binning variables in a way that is meaningful for their research. A user interested in a matching package that does allow for continuous covariates that is still in the Almost Matching Exactly framework may consider exploring \pkg{R-MALTS} or \pkg{pymalts}. These packages implement the algorithm Matching After Learning To Stretch (MALTS), which will use exact matching for discrete variables, and will learn Mahalanobis distances for continuous variables. Instead of a predetermined distance metric like \pkg{MatchIt}, MALTS gives covariates that contribute more towards predicting the outcome higher weights \citep{MALTS}.


\section{Examples}\label{sec:examples}

\subsection{Basic example}
Here we offer an example to illustrate API usage, using a simple, small, 4 unit and 4 covariate simulated dataset to illustrate matched groups easily. An example focused on analysis using a real dataset and its corresponding replication is discussed in Section \ref{sec:real-example}. All classes, functions, and parameters used here, as well as additional options for parameters are defined and discussed in Section \ref{sec:user-guide}.

The first step is importing the package. We show the dataframe used here as well. The Pandas dataframe places units in rows and covariates in columns, and requires a column with a boolean variable indicating treatment, and a column for the outcome variable. 

\begin{samepage}
\begin{Sinput}
import pandas as pd
import dame_flame

df = pd.DataFrame([[0,1,1,1,0,5.1], [0,0,1,0,0,5.11], [1,0,1,1,1,6.5],
        [1,1,1,1,1,6.]], 
        columns=["x1", "x2", "x3", "x4", "treated", "outcome"])

print(df.head())

\end{Sinput}
\end{samepage}
\begin{samepage}
\begin{Soutput}
  x1  x2  x3  x4  treated      outcome
0  0   1   1   1           0        5.10
1  0   0   1   0           0        5.11
2  1   0   1   1           1        6.50
3  1   1   1   1           1        6.0
\end{Soutput}
\end{samepage}

The first step in the matching procedure is instantiating a \code{matching} object, with optional parameters that can specify the early stopping criteria, missing data handling methodology, output style, or the machine learning method used to compute the predictive error. All optional parameters are described in more detail in Section \ref{sec:user-guide}. Here, we choose the default options, which includes no missing data handling, no early stopping procedures, and computes predictive error with ridge regression. We choose these options because this dataset does not have any missing data that needs to be handled and because it is a small example, we do not need to stop the algorithm early.

\begin{Sinput}
model = dame_flame.matching.DAME()
\end{Sinput}

The next step is to call the \code{fit} method on the \code{matching} object created above. Here, users must provide a file location of the holdout training dataset, a Pandas dataframe, or a fraction of the input dataset to use for matching, in the parameter \code{holdout_data}.

Additionally, the name of the treatment column and the name of the outcome column can be provided.
\begin{Sinput}
model.fit(df, "treated", "outcome")
\end{Sinput}

At this point, simply calling the code{predict} method with the input dataset produces matched results. 
The return value from the \code{predict} command contains an output table, which consists of the units that were matched to at least one other unit. For each unit that was matched, the table indicates which of the covariates were used for matching, and the covariate values that each unit was matched on. The covariates that were not used to  match the unit are denoted with ``\code{*}'' as their value.

\begin{Sinput}
result = model.predict(df)
print(result)
\end{Sinput}

\begin{Soutput}
   x1   x2   x3   x4
0   *   1    1    1     
1   *   0    1    *     
2   *   0    1    *     
3   *   1    1    1   
\end{Soutput}
Various result summaries are available, including a printout of all matched groups, and the units belonging to each group.  The result of the \code{predict} function, shown above, can also be retrieved by using the following attribute \code{df_units_and_covars_matched} of the matching class. The \code{units_per_group} attribute of the matching class provides an array of arrays. Each sub-array is a matched group, and each item in each sub-array is an int, indicating the unit in that matched group. If matching is done with the parameter \code{repeats=False} when defining the matching class, then no unit will appear more than once. If \code{repeats=True} then the first group in which a unit appears is its main matched group.
\begin{Sinput}
print(model.units_per_group)
\end{Sinput}
\begin{Soutput}
[[0, 3], [1,2]]
\end{Soutput}

This shows us that unit 0 and unit 3 are in a matched group, and that unit 1 and unit 2 are in another matched group.



The \code{utils} functions offer post-processing. In these functions, users must pass as parameters the matching object declared earlier, and for many of the functions, users must pass in a \code{unit_ids} parameter, which can be a single unit or a list of unit ids.

The function that provides matched groups of each unit is \code{MG}. If one unit id was provided, this is a single dataframe containing the main matched group of the unit. If the unit does not have a match, the return will be \code{numpy.nan}. If multiple unit ids were provided, this will be a list of dataframes with the main matched group of each unit provided. 

\begin{Sinput}
mmg = dame_flame.utils.post_processing.MG(matching_object=model, unit_ids=0)
print(mmg)
\end{Sinput}
\begin{Soutput}
      x1    x2    x3    x4    treated    outcome
 0    *     1     1     1     0          5.1
 3    *     1     1     1     1          6.
\end{Soutput}
This shows the main matched group of unit 0 is unit 3, and that covariates that unit 0 and unit 3 matched on are covariates \code{x2}, \code{x3}, and \code{x4}.

The functions in the \code{utils} library also include treatment effect estimators, as defined in Section \ref{sec:def-estimators}, including an estimate for CATE. If one unit id was provided, the return value will be a single float representing the conditional average treatment effect estimate of the unit. This is equal to the CATE of the group that the unit is in. If the unit does not have a match, the return will be \code{numpy.nan}. If multiple unit ids were provided, the return value will be a list of floats with the CATE estimate of each unit provided.

\begin{samepage}
\begin{Sinput}
cate = dame_flame.utils.post_processing.CATE(matching_object=model, unit_ids=0)
print(round(cate,3))
\end{Sinput}
\begin{Soutput}
0.9
\end{Soutput}
\end{samepage}

The \code{ATE} function, to get the ATE estimate only requires a matching object, but does not require a unit id, and returns a float.

\begin{Sinput}
ate = dame_flame.utils.post_processing.ATE(matching_object=model)
print(round(ate,3))
\end{Sinput}

\begin{Soutput}
1.145
\end{Soutput}

As discussed in Section \ref{sec:def-estimators}, the package also offers a second ATE estimator with a corresponding variance estimator. Again, the required parameter is the matching object used earlier.

\begin{samepage}
\begin{Sinput}
var, ate = dame_flame.utils.post_processing.var_ATE(matching_object=model)
print(round(ate))
print(round(var))
\end{Sinput}
\begin{Soutput}
1.145
0.03
\end{Soutput}
\end{samepage}
As is expected, we see that this ATE estimate is the same as or close to the ATE estimate from the other \code{ATE} function.
\subsection{Example analysis} \label{sec:real-example}

 \pkg{dame-flame} is an interpretable matching package because it allows users to quickly and easily understand which covariates were selected to be important for causal inference. This can be useful for practitioners in determining who benefits from treatment the most and where resources should be spent for future treatment. The package also allows users to view various other aspects of the matching process such as the stopping criteria as they use the package. 


Here we demonstrate an experimental use-case for the DAME and FLAME algorithms on the Tennessee's student teachers achievement ratio (STAR) Dataset. This dataset originates from an experiment beginning in 1985, in which elementary school students and their teachers across 79 schools in Tennessee were randomly assigned to classes of small or regular sizes from Kindergarten through 3rd grade \citep{star}. Although data is available for students not participating in the experiment, we limit to the experimental dataset in which treatment was randomized. The results showed that being placed in a small class size led to higher standardized test performance, and long term benefits in increased college entrance exam taking, especially among minority students \citep{kreuger}. 

Our cleaned dataset has around 5000 students. We use a transformed outcome variable by first computing an empirical CDF of math and reading scores for students in regular classes, then computing percentiles of math and reading scores for all students according to this distribution, and finally averaging these two. Our covariates include children's characteristics, teacher's characteristics and school characteristics. The children's characteristics are gender, race (binary, with White and Asian in one group, and all other races in the other group), free lunch status (students who received free lunch at any point in Kindergarten through 3rd grade are in one group), and age in months (binned into deciles). The teacher characteristics include race, gender, and having a higher degree than bachelors. The school characteristics are urbanicity (rural, urban, suburban, and inner city) and a school identification number, with one for each of the 79 schools. 

First, we ensure that using DAME and FLAME on this dataset is appropriate, by ensuring that there is a lack of sensitivity to the train/test split, which we take to be a random 80\%/20\% split. We do so by ensuring that the algorithm matches a sufficient number of units in each case and that the aggregate treatment effect estimates are reasonable. We first run four different trials of DAME on random splits.  As we iterate over our four trials, we save the matching class objects for further analysis of treatment effects. When we declare an object of the matching class, we use the early stopping criteria of stopping when there is a significant increase of $\mathtt{PE}$ from the baseline $\mathtt{PE}$ computed in the first iteration. Then, we run the \code{fit} and \code{predict} methods on the matching class to run the match, as follows:

\begin{samepage}
\begin{Sinput}
models = []
random_seeds = [1111, 2222, 3333, 4444]
for i in range(len(random_seeds)):
    matching_df, holdout_df = train_test_split(df_trunc, test_size=0.2,
         random_state=random_seeds[i])
    model_dame = dame_flame.matching.DAME(repeats=False, verbose=0,
         adaptive_weights='decisiontree',
         early_stop_pe=0.33)
    model_dame.fit(holdout_data=holdout_df, outcome_column_name='outcome')
    model_dame.predict(matching_df)
    models.append(model_dame)
\end{Sinput}
\end{samepage}
The output is:
\begin{Soutput}
Matching stopped while attempting iteration 28 due to the PE fraction early
stopping criterion.
	Predictive error of covariate set would have been 92.12026755312186
Matching stopped while attempting iteration 20 due to the PE fraction early
stopping criterion.
	Predictive error of covariate set would have been 105.55615864739742
Matching stopped while attempting iteration 16 due to the PE fraction early
stopping criterion.
	Predictive error of covariate set would have been 126.2251752950491
Matching stopped while attempting iteration 20 due to the PE fraction early
stopping criterion.
	Predictive error of covariate set would have been 64.27750359184846
\end{Soutput}


Next, we compute the ATE estimate on each of the trials. We do this by iterating over the matching class objects declared above, and calling the \code{utils} function \code{var\_ATE}, which is defined further in Section \ref{sec:user-guide}.
\begin{Sinput}
for i in range(len(models)):
    var, ate = dame_flame.utils.post_processing.var_ATE(models[i])
    print("Trial", i, "matched", len(models[i].df_units_and_covars_matched),
         "units with an ATE of", round(ate,2), "and a variance of ATE of",
          round(var,2))
\end{Sinput}
\begin{Soutput}
Trial 0 matched 1882 units with an ATE of 5.11 and a variance of ATE of 1.35
Trial 1 matched 1904 units with an ATE of 5.95 and a variance of ATE of 1.34
Trial 2 matched 1814 units with an ATE of 4.98 and a variance of ATE of 1.44
Trial 3 matched 1790 units with an ATE of 5.3 and a variance of ATE of 1.41
\end{Soutput}

It is a good sign that each of the random test/train partitions behaved similarly in that  they all matched a similar number of units and have similar ATE estimates. This indicates that the conclusions are not sensitive to the random partitioning. The ATE estimate and its variance estimate indicate that placing students in smaller class sizes caused those students to achieve higher kindergarten test scores by a few percentile points. We conclude this is a reasonable ATE estimate because in the analysis done in \cite{kreuger}, the estimate for the impact of small class sizes provided by Figure 1 in their work, which is based on a linear regression on many similar covariates, is  between  5 and 6 for average percentile of math and reading scores.

Next, we consider whether DAME or FLAME is a better matching method for this dataset. We run four trials of FLAME with the same random test/train split. Again, we define an object of the matching class, this time of the FLAME subclass, and we run the \code{fit} and \code{predict} methods.

\begin{Sinput}
flame_models = []
random_seeds = [1111, 2222, 3333, 4444]
for i in range(4):
    matching_df, holdout_df = train_test_split(df_trunc, test_size=0.2,
         random_state=random_seeds[i])
    model_flame = dame_flame.matching.FLAME(
        repeats=False, verbose=3, adaptive_weights='decisiontree',
        early_stop_pe=0.33)
    model_flame.fit(holdout_data=holdout_df, outcome_column_name='outcome')
    result_flame = model_flame.predict(matching_df)
    flame_models.append(model_flame)
\end{Sinput}

We omit the output of this matching result, which is available with the full replication code on our GitHub. We observe that the FLAME algorithm iterates fewer times than the DAME algorithm and that both stopped iterating according to the stopping criterion \code{early_stop_pe}. Using the \code{var_ATE} function on the FLAME matching objects yields similar estimates as when using DAME.

We summarize the dropping criteria from the full verbose output of FLAME, generated using the parameter \code{verbose=3} in Table \ref{tab:droporder}.  Each entry is a iteration of the algorithm, listing the covariate dropped in that iteration, and the number of units matched after that covariate is dropped. The fact that teacher gender is dropped first in each trial makes sense, since all teachers are female in the cleaned data set. In \cite{kreuger}, teacher gender and teacher race covariates were not used in their linear regression estimating the impact of small class sizes on percentile test scores for kindergarten students.

\begin{figure}[h!]
  \centering
  \includegraphics[width=1.1\textwidth]{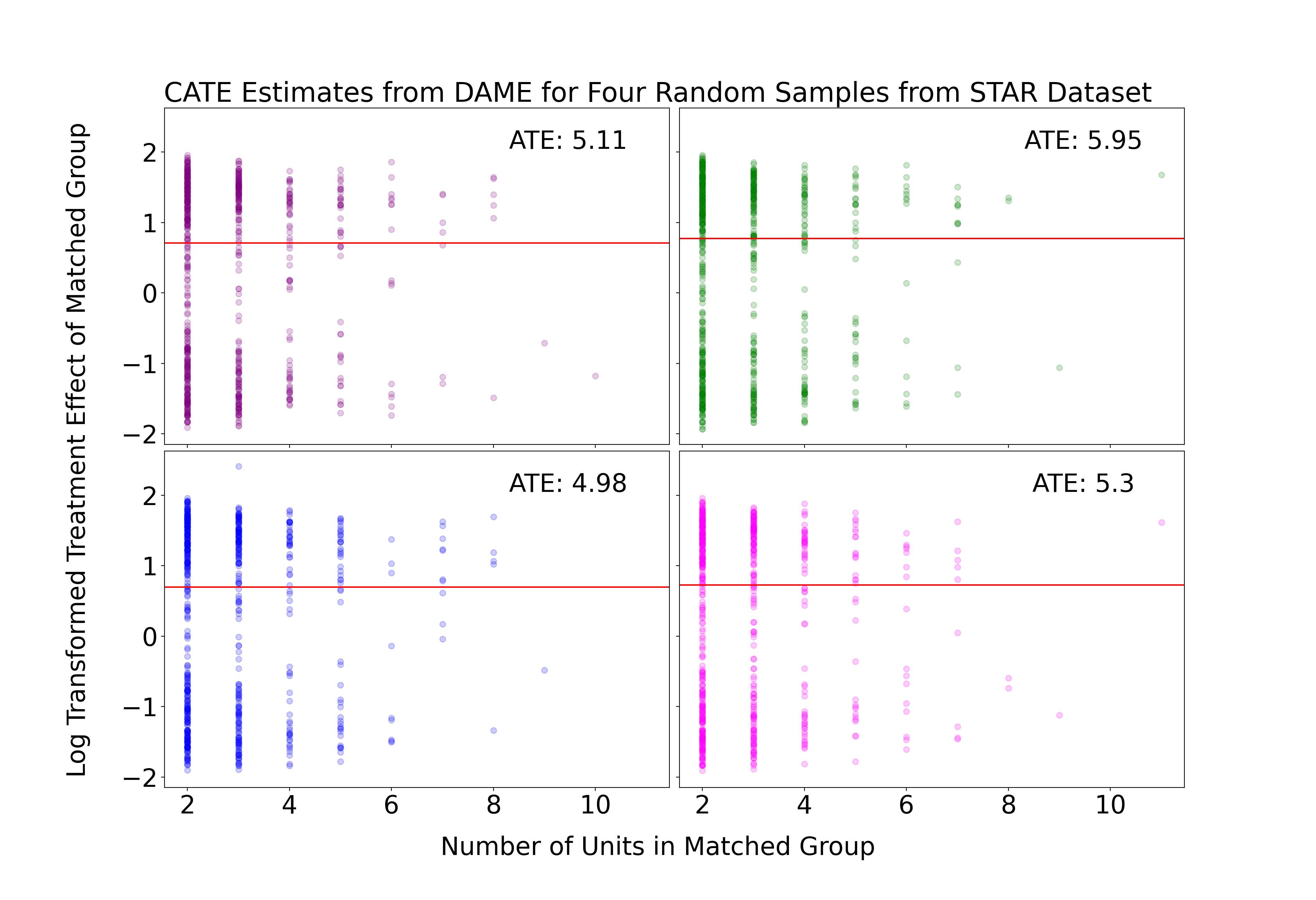}
  \caption{CATE estimates of matched groups, across four random splits of the dataset into matching and holdout training datasets. The vertical line is the ATE estimate.}
  \label{fig:cate-graph6}
\end{figure}

We plot the log transform of CATE estimates of the matched groups from DAME against the number of units of each group in Figure \ref{fig:cate-graph6}. We omit the code to create this plot (available on the GitHub repository) as it highlights \proglang{Python} semantics not specific to \pkg{dame-flame}. This plot highlights heterogeneity in the treatment effects of groups. 

Across the trials, when we examine the matched groups that correspond to the largest group sizes, or the rightmost points on the graphs, we notice that the large matched group contains different units in each trial, so it is sensitive to the test/train split. 

Overall, because the ATE estimates using the DAME and the FLAME algorithms correspond to previous findings, \pkg{dame-flame} proves itself in this situation to be a valuable robustness check for researchers who wish to verify measurable impacts of this well known experiment. We hope that \pkg{dame-flame} can additionally be of use to researchers interested in matching and exploring matched groups and CATE estimates of matched groups.

\begin{table}
\begin{center}

\begin{tabular}{ |c|c|c|c| } 
 \hline
  Trial 1 & Trial 2 & Trial 3 & Trial 4 \\ [0.5ex] 
\hline
 All covariates, 1079 & All covariates, 1053 & All covariates, 1047 & All covariates, 1056 \\ 
 Teacher gender, 0 & Teacher gender, 0 & Teacher gender, 0 & Teacher gender, 0 \\ 
 Urbanicity, 0 & Urbanicity, 0 & Urbanicity, 0  & Urbanicity, 0\\ 
 Teacher race, 198 & Student race, 53 & Teacher race, 209 & Student race, 54 \\ 
Student race, 269 & Teacher race, 265 & Student race, 282  & Teacher race, 252 \\ 
 \hline

\end{tabular}
\caption{
Summary of covariate dropping in order from FLAME.}
\vspace{-3mm}
\label{tab:droporder}
\end{center}
\end{table}

\section{API documentation}  \label{sec:user-guide}
In this section, we provide details on the matching class definitions and function parameters. The API consists of standard Pythonic design established by \pkg{scikit-learn} \citep{sklearn_api}. To begin matching, the user declares an object of type \code{DAME} or type \code{FLAME}. Both of these inherit from the base class \code{MatchingParent}. Following this, the user calls the method \code{fit}, providing a holdout training dataset, and finally \code{predict}, where a user provides a matching dataset, and the matches of the algorithm are computed.

The user provides the input data as a data frame or file in the function \code{predict}, which must contain an outcome column, and a treatment column. FLAME and DAME by default produce an output table consisting of the units that were matched to at least one other unit. For each unit that was matched, the table indicates which of the covariates were used for matching, and the covariate values that each unit was matched on. The covariates that were not used to  match the unit are denoted with \code{"*"} as their values.

\paragraph{Class definition}
\begin{samepage}
\begin{Sinput}
dame_flame.matching.DAME(adaptive_weights='ridge', alpha=0.1, 
         repeats=True, verbose=2, early_stop_iterations=float('inf'), 
         stop_unmatched_c=False, early_stop_un_c_frac=False, 
         stop_unmatched_t=False, early_stop_un_t_frac=False, 
         early_stop_pe=0.05, missing_indicator=numpy.nan, 
         missing_data_replace=0, missing_holdout_replace=0, 
         missing_holdout_imputations=10, 
         missing_data_imputations=1, want_pe=False, want_bf=False)  
\end{Sinput}
\begin{Sinput}
dame_flame.matching.FLAME(adaptive_weights='ridge', alpha=0.1, 
         repeats=True, verbose=2, early_stop_iterations=float('inf'), 
         stop_unmatched_c=False, early_stop_un_c_frac=False, 
         stop_unmatched_t=False, early_stop_un_t_frac=False, 
         early_stop_pe=0.05, missing_indicator=numpy.nan,
         missing_data_replace=0, missing_holdout_replace=0, 
         missing_holdout_imputations=10, 
         missing_data_imputations=1, want_pe=False, want_bf=False)  
\end{Sinput}
\end{samepage}

\pagebreak
Arguments:
\begin{itemize}
    \item \code{adaptive_weights}, type \{bool, \code{`ridge'}, \code{`decisiontree'}, \code{`ridgeCV'},\\\code{`decisiontreeCV'}\}, default=\code{`ridge'}. The method used to decide what covariate set should be dropped next.
    \item \code{alpha}, type float, default=\code{0.1}. If \code{adaptive_weights} is set to \code{`ridge'}, this is the alpha for ridge regression.
    \item \code{repeats}, type bool, default=\code{True}. Whether or not units for whom a main matched has been found can be used again, and placed in an auxiliary matched group.
    \item \code{verbose}, type int: \{\code{0}, \code{1}, \code{2}, \code{3}\}, default=\code{2}. Style of printout while algorithm runs. If \code{0}, no output. If \code{1}, provides iteration number. If \code{2}, provides iteration number and additional information on the progress of the matching at every 10th iteration. If \code{3}, provides iteration number and additional information on the progress of the matching at every iteration.
    \item \code{early_stop_iterations}, type \{float, int\}, default=\code{float(`inf')}. A number of iterations after which to hard stop the algorithm.
    \item \code{stop_unmatched_c}, type bool, default=\code{False}. If \code{True}, then the algorithm terminates when there are no more control units to match.
    \item \code{stop_unmatched_t}, type bool, default=\code{False}. If \code{True}, then the algorithm terminates when there are no more treatment units to match.
    \item \code{early_stop_un_c_frac}, type float, default=\code{0.1}. Must be between 0.0 and 1.0. This provides a fraction of unmatched control units. When the threshold is met, the algorithm will stop iterating. For example, using an input dataset with 100 control units, the algorithm will stop when 10 control units are unmatched and 90 are matched (or earlier, depending on other stopping conditions).
    \item \code{early_stop_un_t_frac}, type float, default=\code{0.1}. Must be between 0.0 and 1.0. This provides a fraction of unmatched treatment units. When the threshold is met, the algorithm will stop iterating. For example, using an input dataset with 100 treatment units, the algorithm will stop when 10 control units are unmatched and 90 are matched (or earlier, depending on other stopping conditions).
    \item \code{early_stop_pe}, type float, default=\code{0.05}. If the algorithm attempts to drop a covariate set that would lead to a predictive error above (1 + \code{early_stop_pe}) times the baseline predictive error (the predictive error when using all covariates to predict), then the algorithm terminates before dropping this covariate set.
    \item \code{want_pe}, type bool, default=\code{False}. If \code{True}, the output of the algorithm will include the predictive error of the covariate sets used for matching in each iteration.
    \item \code{want_bf}, type bool, default=\code{False}. If \code{True}, the output will include the balancing factor for each iteration.
    \item \code{missing_indicator}, type \{character, integer, \code{numpy.nan}\}, default=\code{numpy.nan}. This is the indicator for missing data in the dataset.
    \item \code{missing_holdout_replace}, type int: \{\code{0}, \code{1}, \code{2}\}, default=\code{0}. If \code{0}, assume no missing holdout training data and proceed. If \code{1}, the algorithm excludes units with missing values from the holdout dataset. If \code{2}, do MICE on holdout dataset. If this option is selected, it will be done for a number of iterations equal to \code{missing_holdout_imputations}.
    \item \code{missing_data_replace}, type int: \{\code{0}, \code{1}, \code{2}, \code{3}\}, default=\code{0}. If \code{0}, assume no missing data in matching data and proceed. If \code{1}, the algorithm does not match on units that have missing values. If \code{2}, prevent all \code{missing_indicator} values from being matched on. If \code{3}, do MICE on matching dataset. This is not recommended. If this option is selected, it will be done for a number of iterations equal to \code{missing_data_imputations}.
    \item \code{missing_holdout_imputations}, type int, default=\code{10}. If \code{missing_holdout_replace=2}, this is the number of imputations.
    \item \code{missing_data_imputations}, type int, default=\code{1}. If \code{missing_data_replace=3}, this is the number of imputations.
\end{itemize} 

\paragraph{\code{fit} function}

\begin{Sinput}
fit(self, holdout_data=False, treatment_column_name=`treated', 
    outcome_column_name=`outcome', weight_array=False))

\end{Sinput}

Arguments:
\begin{itemize}
    \item \code{holdout_data}, type \{string, dataframe, float, \code{False}\}, default=\code{False}. This is the holdout training dataset. If a string is given, that should be the location of a CSV file to input. If a float between 0.0 and 1.0 is given, that corresponds the percent of the input dataset to randomly select for holdout data.  If \code{False}, the holdout data is equal to the entire input data. If users choose to use units repeatedly in both the holdout and training dataset, they should be careful that the data do not have a special situation that need to be respected in subsampling such as a hierarchy.
    \item \code{treatment_column_name}, type string, default=\code{"treated"}. This is the name of the column with a binary indicator for whether a row is a treatment or control unit.
    \item \code{outcome_column_name}, type string, default=\code{"outcome"}. This is the name of the column with the outcome variable of each unit.
    \item \code{weight_array}, type array, optional. If \code{adaptive_weights=False}, these are the weights to the covariates in \code{input_data}, for the non adaptive version of DAME. Must sum to 1. In this case, we do not use machine learning for the weights, they are manually entered as \code{weight_array}.
\end{itemize}

\paragraph{\code{predict} function}

\begin{Sinput}
predict(self, input_data)
\end{Sinput}

Argument for both FLAME and DAME objects:
\begin{itemize}
    \item \code{input_data}, type \{string, dataframe\}, Required Parameter. The dataframe on which to perform the matching, or the location of the CSV with the dataframe. 
\end{itemize}

Arguments for FLAME object only:
\begin{itemize}
    \item \code{C}, type float, default=\code{0.1}. The tradeoff parameter between the balancing factor and the predictive error when deciding which covariates to match on.
    \item \code{pre_dame}, type {float, integer}, default=\code{float(`inf')}. The number of iterations to run FLAME before switching to DAME, allowing for a hybrid FLAME-DAME option.
\end{itemize}

Return values:
\begin{itemize}
    \item \code{Result}. Pandas dataframe of matched units and covariates matched on, with a \code{"*"} at each covariate that a unit did not use in matching.                     
\end{itemize}

\paragraph{Matching class attributes}
\begin{itemize}
    \item \code{units_per_group}, type array. This is an array of arrays. Each sub-array is a matched group, and each item in each sub-array is an int, indicating the unit in that matched group. If matching is done with \code{repeats=False} when defining the \code{DAME} or \code{FLAME} matching object, then no unit will appear more than once. If \code{repeats=True} then the first group in which a unit appears is its main matched group.
    \item \code{df_units_and_covars_matched}, type dataframe. This is the resulting matches of DAME. Each matched unit is in this array, and the covariates they were matched on have the value used to match. The covariates units were not matched on are indicated with a \code{"*"}.
    \item \code{groups_per_unit}, type array. The length of this is equal to the number of units in the input array. Each item in this array corresponds to the number of times that each item was matched. If matching is done with \code{repeats=False} when defining the matching objects, then this number will be either \code{0} or \code{1}.
    \item \code{bf_each_iter}, type array. If argument \code{want_bf=True} when defining the matching class, this will contain the balancing factor of the chosen covariate set at each iteration.
    \item \code{pe_each_iter}, type array. If argument \code{want_pe=True} when defining the matching class, this will contain the predictive error of the chosen covariate set at each iteration.
\end{itemize}

\paragraph{Post-processing utils}

\begin{Sinput}
dame_flame.utils.post_processing.MG(matching_object, unit_ids, output_style=1, 
     mice_iter=0)
dame_flame.utils.post_processing.all_MGs(matching_object)
dame_flame.utils.post_processing.ATE(matching_object, mice_iter=0)
dame_flame.utils.post_processing.CATE(matching_object, unit_ids, mice_iter=0)
dame_flame.utils.post_processing.ATT(matching_object, mice_iter=0)
dame_flame.utils.post_processing.var_ATE(matching_object)
\end{Sinput}

Arguments:
\begin{itemize}
    \item \code{matching_object}, type {\code{dame_flame.matching.DAME}, \code{dame_flame.matching.FLAME}}. The matching object used to run DAME and FLAME, after the \code{fit} and \code{predict} methods have been called to create the matches. If the \code{matching_object}'s parameter for \code{verbose} is not \code{0}, then as units without matches appear, the function will print that those units did not have matches.
    \item \code{unit_ids}, type \{int, list\}. A unit id or list of unit ids. 
    \item \code{output_style}, type int: \{\code{0},\code{1}\}. Default=\code{1}. If \code{1}, the covariates which were not used in matching for the group of the unit will have a \code{"*"} rather than the covariate value. Otherwise, it will output all covariate values.
    \item \code{mice_iter}, type int. Default=\code{0}. If matching was done using MICE, this is the iteration of MICE for which the treatment effects or matched groups will be found.
\end{itemize}

Return values and corresponding functions:

\begin{itemize}
    \item \code{MG}: type \{list, dataframe, \code{numpy.nan}\}. Returns matched groups corresponding to inputs unit ids. If one unit id was provided, this is a single dataframe containing the main matched group of the unit. If the unit does not have a match, the return will be \code{numpy.nan}. If multiple unit ids were provided, this will be a list of dataframes with the main matched group of each unit provided. If any unit does not have a match, rather than a dataframe, at its place will be \code{numpy.nan}.
    \item \code{all_MGs}: type {dictionary}. Returns all matched groups, in a dictionary mapping a unit id that was matched to a list of unit ids that are in the main matched group of that unit. 
    \item \code{ATE}: type \{float, \code{numpy.nan}\}. Returns the average treatment effect (ATE) estimate in a float. If no units were matched, then the output will be \code{numpy.nan}.
    \item \code{CATE}: type \{list, float, \code{numpy.nan}\}. Returns the conditional average treatment effects (CATE) estimates. If one unit id was provided, this is a single float representing the estimate of the conditional average treatment effect of the unit. This is equal to the CATE estimate of the group that the unit is in. If the unit does not have a match, the return will be \code{numpy.nan}. If multiple unit ids were provided, this will be a list of floats with the CATE of each unit provided. If any unit does not have a match, rather than a float within the list, at its place will be \code{numpy.nan}.
    \item \code{ATT}: type \{float, \code{numpy.nan}\}. Returns the average treatment effect on treated (ATT) estimate in a float. If no units were matched, then the output will be \code{numpy.nan}.
    \item \code{var_ATE}: Tuple with two items. The first is the variance of ATE estimate in type float. The second is the ATE estimate in type float. These estimates are computed using the methodology in \cite{abadie}.

\end{itemize}

\section{Conclusions}
The FLAME and DAME algorithms for matching of observational data with discrete covariates provide interpretable and high-quality matches.
The \pkg{dame-flame} open-source \proglang{Python} package offers efficient, easy-to-use implementations of these algorithms. The package is easily accessible, and here, we provide detailed documentation, with concrete examples. The package is written in a highly modular manner, facilitating the introduction of new features 
and variations of the DAME and FLAME algorithms. It is available at \url{https://github.com/almost-matching-exactly/DAME-FLAME-Python-Package}. We believe this package will be a useful tool for social science researchers, health researchers, and other scientists that use matching. 

%

\section*{Acknowledgments}
This work was supported in part by awards NIH 
R01EB025021, NSF 
IIS-1703431, and
the 
Duke University Energy Initiative Energy Research
Seed Fund. 


\bibliography{refs}

\end{document}